# Exploration of Adolescent Depression Risk Prediction Based on Census Surveys and General Life Issues.

Qiang LI[1], Yufeng Wu[2], Zhan Xu[3], Hefeng Zhou[*2]
[1]Shangrao Normal University
[2]Shanghai Jiao Tong University | zadeji1@sjtu.edu.cn
[3]University of Michigan Ann Arbor

**Abstract:** In contemporary society, the escalating pressures of life and work have propelled psychological disorders to the forefront of modern health concerns, an issue that has been further accentuated by the COVID-19 pandemic. The prevalence of depression among adolescents is steadily increasing, and traditional diagnostic methods, which rely on scales or interviews, prove particularly inadequate for detecting depression in young people. Addressing these challenges, numerous AI-based methods for assisting in the diagnosis of mental health issues have emerged. However, most of these methods center around fundamental issues with scales or use multimodal approaches like facial expression recognition. Diagnosis of depression risk based on everyday habits and behaviors has been limited to small-scale qualitative studies. Our research leverages adolescent census data to predict depression risk, focusing on children's experiences with depression and their daily life situations. We introduced a method for managing severely imbalanced high-dimensional data and an adaptive predictive approach tailored to data structure characteristics. Furthermore, we proposed a cloud-based architecture for automatic online learning and data updates. This study utilized publicly available NSCH youth census data from 2020 to 2022, encompassing nearly 150,000 data entries. We conducted basic data analyses and predictive experiments, demonstrating significant performance improvements over standard machine learning and deep learning algorithms. This affirmed our data processing method's broad applicability in handling imbalanced medical data. Diverging from typical predictive method research, our study presents a comprehensive architectural solution, considering a wider array of user needs.

## 1. Introduction

With the maturity of computer technology, various intelligent algorithms and industrial solutions have been invented. With the development of information mining technology, big data has been widely used in various industries. At present, computer-aided diagnosis has been used in the early detection detection and intervention in cardiovascular disease[1], diabetes mellitus and cancer have led to substantial reductions in morbidity and mortality and improved quality of life among individuals with these conditions. In the face of the fast-paced society, mental illness has increasingly become a contemporary factor affecting health. Teenagers are in the second stage of resistance. The incidence of psychological problems of teenagers is 15% - 32%, and it is on the rise[2]. The psychological problems of teenagers can not be ignored. Common psychological problems of adolescents include interpersonal relationship problems, emotional stability problems and learning emotional problems

With the development of society and medical technology, mental illness has become an important part of medical field. Due to the relatively immature physical and psychological development of teenagers, if there are various adverse factors in the growing environment, such as apathy of parent-child relationship, lack of warm dependency, tense family atmosphere, coarse educational methods, and suppressed peer relationship, it is possible to cause teenagers to have problem behaviors. To carry out the research on the psychological correction strategies of teenagers' problem behaviors is conducive to the maintenance of their psychological health, the sound development of their individuality and socialization, the comprehensive implementation of national quality education and the promotion of the construction of socialist spiritual civilization.

At present, the efforts to extend the general physiological risk detection method to schizophrenia have been focused on developing and verifying standards to identify the risk individuals (i.e. clinically high-risk or prodromal patients) with impending onset of psychosis, and to track the changes of the disease over time.[3,4] compared with the high-risk method based on the family history of schizophrenia, one advantage of this method is that the assessment related to the disease onset can more effectively identify potential psychological problems and carry out intervention treatment in advance.

The research on big data mental health in foreign countries is at the forefront level. At present, there are related topics on monitoring the occurrence of mental health problems by using big data technology. Five years ago, the United States started the project of the survey database on psychological problems of adolescents. The purpose of this project is to study the psychological state of adolescents, especially students, and hope to design a questionnaire data screening judgment methodology to assist in the diagnosis of adolescent mental diseases.In the project, we designed a potential psychological problem risk prediction method based on a questionnaire survey according to expectations. The analysis results were obtained through data collection, data processing, and neural network calculations. The data is sourced from the American Association of Adolescent Health, and we will screen and clean the potentially useful data to obtain the target dataset.

There have been a lot of disease prediction in the past, but it is rarely involved in psychological diagnosis and treatment. Before, we investigated the related decision tree is a commonly used classifier in data mining. For medical data, the classification results are required to be high, and the results can only be used as reference to provide auxiliary diagnosis and treatment: [5] In the UK, the current National Institute for Health and Care Excellence (NICE) guidelines recommend use of the QRISK2 score to guide the initiation of lipid lowering therapies.[6,7]. Existing risk prediction algorithms are usually developed using multiple regression models, which combine information of a limited number of mature risk factors and generally assume that all these factors are related to cardiovascular disease outcomes in a linear manner, with limited or no interaction between different factors. However, at present, only a few studies have investigated the potential advantages of using ML methods for CVD risk prediction, focusing only on a limited number of ML methods [8,9] or a limited number of risk prediction [10].

Although psychological pathology is rarely studied, some explorations have been made in this field. In the application of artificial intelligence in the field of mental health, the most notable cases are used for suicide prevention. Machine learning algorithms can not only isolate and identify suicide risk factors, but also integrate the relationships between research variables and evaluate these variables. In 2016, Passos et al. analyzed the past psychiatric diagnoses, drug use, and treatment history of depression in an attempt to draw conclusions. In the same year, Song et al. used multi-level modeling of internet search patterns to identify risk factors of interest[11]. Research has been conducted on predicting suicide risk, and because suicide assessment is complex and influenced by various factors, traditional statistical analysis methods have limitations in analyzing complex data and have been unable to predict suicide behavior beyond chance levels[12]. In the field of natural language processing, there have been many attempts to detect depression from social media posts. The most widespread feature engineering method uses Language Inquiry Word Count (LIWC), which extracts lexical features from a lexicon of more than 32 types of psychological structures[13].

In this work, we completed the establishment of a youth health dataset by collecting partial information from the national health database. We have developed a data processing tool CensusFilter based on the SMOTE and ENN algorithms tool, and we have also built an Adaptive Adolescent Depression Assistant (AADA) Predictor that can be learned online. The core algorithm of this system is based on the AutoML algorithm. The algorithm core can be deployed in the cloud, providing APIs as a diagnostic basis for adolescent depression. The contribution points of this article are as follows:

- Collected and organized the adolescent health survey data released by NSCH from 2020 to 2022, and created a dataset available for machine learning. In terms of data processing, we propose using data similarity calculation to determine the importance of data samples, and sampling to establish sub datasets based on data importance to improve the problem of severe sample imbalance leading to performance degradation of machine learning models.
- Based on the idea of Automatics Machine Learning, we propose an automatic prediction algorithm that can be learned online. The algorithm can automatically generate new model structures when the training data changes, improving the generalization and usability of the prediction algorithm. In addition, we have proposed a cloud service framework that allows algorithm models to directly serve medical assisted judgment and ordinary user assisted self diagnosis.
- We have created different datasets based on census data to validate our proposed method. The experimental results indicate that our method has a significant effect on the auxiliary diagnosis of adolescent depression. Meanwhile, for the diagnosis of adolescent depression, we have also provided partial cause weights in different modeling methods, which can prevent the early occurrence of adolescent depression by focusing on items with higher weights.

**2. Material and Methods**

In previous studies, there has been significant research on disease prediction, but there is limited use of big data for basic psychological diagnosis. On one hand, psychological disorders are considered relatively latent diseases, making it difficult for conventional pathological detection instruments to identify the underlying changes and establish a diagnosis. On the other hand, there are significant challenges in collecting and processing experimental data.

To address these challenges, we propose a method that aims to achieve data accessibility through simple data collection methods. Additionally, the data preprocessing techniques employed in our approach will be utilized to make the research efficient.

2.1. Data Resource

Through research, we located relevant study data, which at present come from the National Survey of Children's Health (NSCH) in the United States[14]. The NSCH provides rich data on various crossover aspects of children's lives, including physical and mental health, opportunities to access quality medical services, and the family, community, school, and social environments of children. The U.S. Census Bureau conducted revised surveys via mail and online from 2016 to 2022. Among other changes, the 2016 National Survey of Children's Health began integrating two surveys: the previous NSCH and the National Survey of Children with Special Health Care Needs (NS-CSHCN). For more information about the National Survey of Children's Health from 2016 to 2022, methods, survey content, and data availability, please refer to the MCHB website.

In the census data from 2020 to 2022, the entries for each set of data amount to 42777, 50892, and 54103 respectively. According to the Topical Variable List, the item 'hhid' refers to the Unique Household ID, and all of the 'hhid' values in the three years of the survey are different. The data regarding the three years of survey data and the incidence of depression in children are as shown in **Table 1**.

**Table 1.** 2020-2022 Census Database Statistics

| Title | 2020 Topical Data | 2021 Topical Data | 2022 Topical Data |
|---|---|---|---|
| Items | 42777 | 50892 | 54103 |
| Variables | 443 | 463 | 490 |
| Has Been Diagnosed Depression | 2316 | 2605 | 3103 |
|  | 5.40% | 5.10% | 5.70% |
| Currently Depression | 1851 | 2095 | 2498 |
|  | 4.3% | 4.1% | 4.6% |

From the statistics, we can see that among adolescents aged 0-17, approximately 5%-6% have suffered from depression, and the rate of those currently suffering from depression is about 4.5%. The statistical tables also reveal that the number of data variables in the census data from 2020 to 2022 all exceed 400. After aligning the data, we selected all items with a correlation greater than 0.05 to depression through Pearson correlation analysis.

We selected 30 variables for research from five dimensions—Personal Information, Physical Health, Mental Health, Living/Study, and Family—by consulting experts. The specific details are shown in the table. We also calculated the Number of Default Values for each item. In the presented data, all meet statistical norms, and the quantity of default values in all data does not exceed 4.3% for any item. All selected items are statistically valid.

**Table 2.** Selected Variables and data default situation

| Variable Category | Variable | Explaination | Number of Default Values | | |
|---|---|---|---|---|---|
| | | | 2020 | 2021 | 2022 |
| Personal Information | sc_age_year | Age of the Child | - | - | - |
| | docprivate | Child Spoke with Doctor Privately | 186 | 140 | 168 |
| | sc_sex | Sex of Selected Child | - | - | - |
| | overweight | Doctor Identified as Overweight | 173 | 234 | 275 |
| Physical Health | stomach | Difficulty Stomach Past 12 Months | 209 | 383 | 432 |
| | headache | Headaches | 99 | 114 | 166 |
| | toothaches | Difficulty Toothaches Past 12 Months | 223 | 349 | 407 |
| | breathing | Difficulty Breathing Past 12 Months | 80 | 146 | 195 |
| | allergies | Allergies | 63 | 107 | 109 |
| Mental Health | memorycond | Difficulty Concentrating, Remembering, or Decisions | 255 | 195 | 253 |
| | errandalone | Difficulty Doing Errands Alone | 182 | 148 | 199 |
| | k2q32a | Depression | 114 | 166 | 257 |
| | k2q31a | ADD/ADHD | 188 | 283 | 364 |
| | k2q33a | Anxiety | 144 | 198 | 278 |
| | k2q38a | Tourette Syndrome | 151 | 230 | 284 |
| Living/Study | makefriend | Difficulty Making or Keeping Friends | 614 | 577 | 485 |
| | k7q30 | Sports Team or Sports Lessons - Past 12 Months | 494 | 409 | 506 |
| | k7q33 | How Often Attend Events or Activities | 450 | 350 | 487 |
| | hoursleep | How Many Hours of Sleep Average | 568 | 464 | 478 |
| | screentime | How Much Time Spent with TV, Cellphone, PC | 770 | 959 | 990 |
| | physactiv | Exercise, Play Sport, or Physical Activity 1 Hour | 428 | 338 | 455 |
| | bedtime | Go to Bed at Same Time | 538 | 770 | 777 |
| | k7q02r_r | Days Child Missed School - Illness or Injury | 415 | 351 | 447 |
| Family | a1_age | parental age | 1296 | 1607 | 1692 |
| | ace3 | Parent or Guardian Divorced | 1484 | 1887 | 2097 |
| | ace5 | Parent or Guardian Time in Jail | 1577 | 2025 | 2258 |
| | ace6 | Adults Slap, Hit, Kick, Punch Others | 1607 | 2051 | 2306 |
| | ace7 | Victim of Violence | 1612 | 2064 | 2286 |
| | ace8 | Lived with Mentally Ill | 1632 | 2072 | 2313 |
| | ace9 | Lived with Person with Alcohol/Drug Problem | 1630 | 2029 | 2307 |

2.2. Methods

In previous studies, there has been significant research on disease prediction, but there is limited use of big data for basic psychological diagnosis. On one hand, psychological disorders are considered relatively latent diseases, making it difficult for conventional pathological detection instruments to identify the underlying changes and establish a diagnosis. On the other hand, there are significant challenges in collecting and processing experimental data.

To address these challenges, we propose a method that aims to achieve data accessibility through simple data collection methods. Additionally, the data preprocessing techniques employed in our approach will be utilized to make the research efficient.

2.2.1. Cloud-based frameworks

In order to facilitate the use of users, we propose the cloud architecture model of services. At the bottom of the architecture, we support the proposed data enhancement algorithm and automatic classification algorithm, which provide update services for the cloud online model. At the same time, the algorithm we design ensures that the model can be learned online, and the model can be adjusted according to the latest data without offline to achieve the best judgment effect. Considering the privacy of medical data, we added an encryption algorithm between the client side and the cloud model to ensure the security of the data. The encryption module was not the focus in this study, so we only used the basic homomorphic encryption algorithm to design the whole architecture as shown in the figure below.

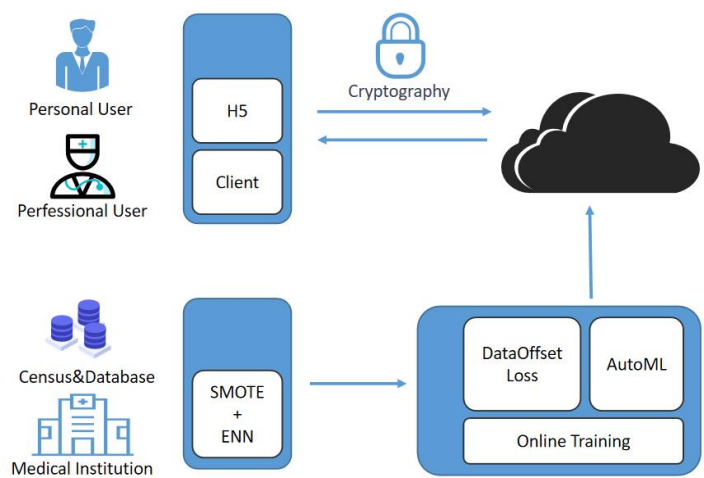

**Figure 1.** The architecture of our system.

The training data source is sourced from censuses and databases. Afterwards, we used our own proposed data processing methods to process this data and create a dataset that the model algorithm can use. After inputting the dataset into our proposed algorithm method, we propose a model structure that adapts to the dataset through the framework of AutoML, and conduct online learning training. After training verification, manually verify and upload updates to the cloud. Different users can access our cloud model through web pages or clients to obtain auxiliary diagnostic results.

2.2.2. Data Process

2.2.2.1. Targeted Data Augmentation Method for Medical Applications

Data augmentation[15] is a commonly used technique in machine learning for increasing the quantity and diversity of limited data to increase the robustness. Data augmentation methods currently include paraphrasing-based, noise-based, sample-based, and other approaches [16]. In the case of a small data collection in medical samples, it is necessary to employ data augmentation methods to expand the dataset while ensuring data accuracy. Typically, natural collection of positive samples in medical data is significantly lower than negative samples, necessitating scientifically supplementing the data. Compared to general data augmentation methods, we propose a targeted data processing approach for medical data:

a) Random Duplication of Positive Samples

By duplicating positive samples, we increase the proportion of positive samples in the dataset. We refer to this approach as "Depression_Duplicated."

b) Adding Noisy Data

Through a seed distribution algorithm, we randomly select a few real positive samples and create noise by duplicating and adjusting the sampling values of a small number of positive samples. This enhances the robustness of the model after training.

c) Data Balancing

It is necessary to balance the data by adjusting the quantity to ensure a balanced distribution of positive and negative samples, removing individuals with completely identical evaluation values. To address the above requirements, we propose using a hybrid method of SMOTE and ENN to manipulate the data.

- Adaptive SMOTE Method: In an imbalanced dataset where the negative samples are significantly fewer than the positive samples, we only need to apply SMOTE to the positive samples. Using our proposed implementation approach, the generated data from SMOTE will be labeled as synthetic data.
- Adaptive ENN Method: Based on the data entry scoring and labeling mechanism we proposed, there is no need to remove the under-sampled data generated by ENN. Instead, we simply decrease its importance coefficient in the database.

In summary, we suggest using a combination of SMOTE and ENN, where SMOTE is applied only to the positive samples, and ENN-generated under-sampled data is labeled and adjusted in the database based on the entry scoring mechanism.

2.2.2.2. database with information evaluation scores and relevant labels

In typical data processing methods, certain samples are often removed to achieve balanced data, which may do harm to the richness of the original data. Since general data augmentation and filtering techniques may not adequately address severe data imbalance issues, we proposed a method CensusFilter to use data with item scores and a data extraction algorithm. In this algorithm, each data entry is assigned a score which represents the importance. The scores range from 0 to 1, with higher values indicating higher importance.

The following formula describes the equation for scoring importance, where $\omega_1$, $\omega_2$ represent positive samples (which are generally fewer in quantity in this study) and negative samples, respectively. The coefficient $\theta_i$ is a manually defined hyperparameter, and we use default values $\theta_1 = 0.9$, $\theta_2 = 1$. The influence of hyperparameters on data sampling will be discussed in subsequent experiments.

$$\omega_1 = \frac{\theta_1}{ln\,(e - 1 + n_k)}$$

$$\omega_2 = \frac{n_p}{n_n} \cdot \frac{\theta_2}{ln\,(e - 1 + n_k)}$$

where $n_k$ represents the number of data entries in a similar class.

In this study, due to the high dimensionality and large volume of the data, using conventional clustering algorithms may result in excessively large clusters. Therefore, we choose the DBSCAN method [17] to implement the clustering functionality for different items. To handle high-dimensional features of the data source, we select the HDBSCAN [18] implementation. HDBSCAN (Hierarchical Density-Based Spatial Clustering of Applications with Noise) is a density-based clustering algorithm. The "Hierarchical" in its name reflects a key characteristic of this algorithm, which reflects the density between data points and the hierarchical structure of these densities, allowing for the discovery of clusters with different densities. It is also necessary to consider the diversity of data categories; if the data clusters are too large, the sampling coefficient will be reduced. Therefore, we improve the HDBSCAN algorithm by adding a constraint on the maximum number of clusters. We set the parameters min_samples = 3 and max_samples = 20 to ensure the diversity of sample categories and to prevent excessively low sampling probabilities for samples in the same category. Through calculations, when $\theta_1 = 1.35$ is used, the minimum importance $\omega_1 = 0.4$.

Meanwhile, in our work, we differentiate between original data and generated data by assigning labels which is 0 or 1. It indicate whether the generated item is from the original data. Then we have the final formula for the label score.

$$\omega_{gen} = \begin{cases} \theta_{gen}, n_{gen} \leq n_p \\ \dfrac{\theta_{gen} n_{gen}}{n_p}, n_{gen} > n_p \end{cases}$$

Where $\theta_{gen}$ is evaluated manually, from 0.1 to 0.2. $N_p$ is the number of positive samples.

2.2.3. Prediction Algorithm

2.2.3.1. Traditional Machine Learning Method

Before the appearance of deep learning, traditional machine learning methods played a leading role in the field of artificial intelligence. These methods utilized statistical techniques to enable computer programs to learn patterns from data and make predictions on new data. Logistic regression and XGBoost are commonly used machine learning methods for classification problems. In our comparative experiments, we will select the simplest logistic regression method as the baseline model.

2.2.3.2. Deep learning prediction methods

One prominent architecture in deep learning is the multilayer perceptron (MLP), which has made deep learning feasible. In our research, we propose an innovative algorithm and use a simple MLP as a comparative model. The MLP consists of multiple layers, including an input layer, hidden layers, and an output layer. We set the number of hidden layers to 3 in this case. Each layer contains 5-20 neurons, which are interconnected through fully connected connections. Additionally, we apply dropout to enhance the model's robustness.

2.2.3.3. Sampling Combination Prediction Method based on AutoML

In previous risk prediction research, machine learning methods such as regression, decision trees, XGBoost, and SVM were commonly used for datasets with multiple data distributions. These methods were suitable because the dimensions and data were not complex, making them easy to handle. However, for our datasets, which has a large volume of data and multiple analysis items, it is necessary to customize the prediction models based on the data because manual modeling will be time-consuming and labor-intensive. To solve this, we introduce the concept of AutoML for automated prediction model building. AutoML proposes the idea of adjusting the model based on the feedback of loss obtained during training on different datasets. Model adjustments are performed within a predefined search space. Such search actions are typically divided into two categories: model structure search[19,20] and hyperparameter search[21,22].

We choose the mainstream method in AutoML methods, Bayesian optimization[23]. As an example, the Auto-Weka series [24] and Auto-Sklearn series [25] employ Bayesian optimization to select combinations and parameters, achieving better performance on datasets. However, considering the limitations of AutoML in handling large-scale data, we draw inspiration from Boosting and Rainbow methods and perform multiple sampling on the scored label data to create different datasets. These datasets are then used to build multiple models, which are combined into an Advanced Predictor for prediction purposes.

In order to improve management, we have also enhanced the loss calculation method, taking inspiration from the Focal Loss calculation approach for imbalanced data. Considering binary classification and cross-entropy loss:

$$CE = -(y\log(p) + (1-y)\log(1-p)) = \begin{cases} -\log(p), & y = 1 \\ -\log(1-p), & y = 0 \end{cases}$$

Building upon cross-entropy loss, we introduce a weighting factor α to reduce the impact of positive samples on the loss when the predicted value p for positive samples is greater than 0.5. Conversely, we perform the opposite operation when p is less than 0.5, setting α=1-p. This extends Focal Loss to the multi-class scenario:

$$FL(p_t) = -\alpha_t(1-p_t)^\gamma \log(p_t)$$

In our implementation of automated machine learning, we also incorporate data volume weighting as follows, where N represents the total sample size in the current data extraction and n_t represents the data volume for that class. This design significantly improves the balance of the calculations with respect to the samples.

$$FL'(p_t) = -\alpha_t(1-p_t)^{\frac{1}{2}}\left(\frac{n_t}{N}\right)^{\frac{1}{2}}\log(p_t)$$

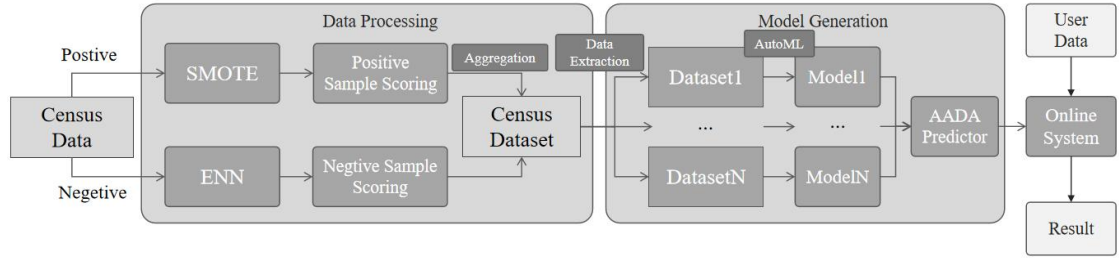

**Figure 2.** The workflow of the entire system contains of data process and modeling. The original data is divided into positive and negative sample, while different meta methods are employed for different categories. After aggregation, different batch subsets are sampled by extraction. Adaptive modeling method for train subsets generates diverse single model predictor, which proposes for the boosted predictor (AADA Predictor).

D. Attention Mechanism

The attention mechanism is a weighted allocation mechanism based on the relevance of factors in the face of multiple potential influences. It selects the most relevant tuples for input into the analysis model. We utilize attention mechanisms [26] to improve model accuracy and enhance computational speed when dealing with large-scale and high-dimensional input data。

## 3. Results

### 3.1. Experiment Data and Analysis

In our methodology, we proposed data processing and prediction methods. At the same time, we also listed ordinary machine learning methods and simple deep network methods as comparative methods for prediction, and the prediction results are shown in the table. To avoid the impact of model bias caused by negative samples exceeding 95% of the entire dataset, we also established a balanced dataset for testing. The most basic split ratio for the training set and test set is 8:2.

We also conducted basic data analysis on the original data, such as correlation analysis and feature importance analysis, to demonstrate the usability of the data.

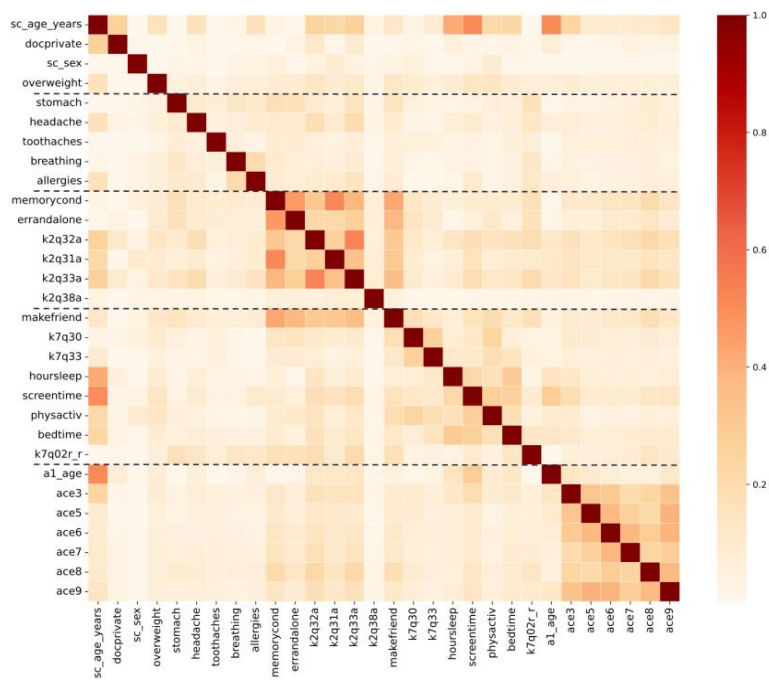

**Figure 3.** The heatmap displays the correlation between entries from different dimensions. A black dashed line in the figure separates different major categories. There is a high correlation between psychological issues and intra group elements of the living environment. The correlation distribution of some items related to depression (k2q32) is relatively uniform, except for the Mental Health. In addition to the high correlation in the psychological category, the lack of significant correlation in other items highlights the complexity of the original data and the necessity for reasonable data processing methods to decode it.

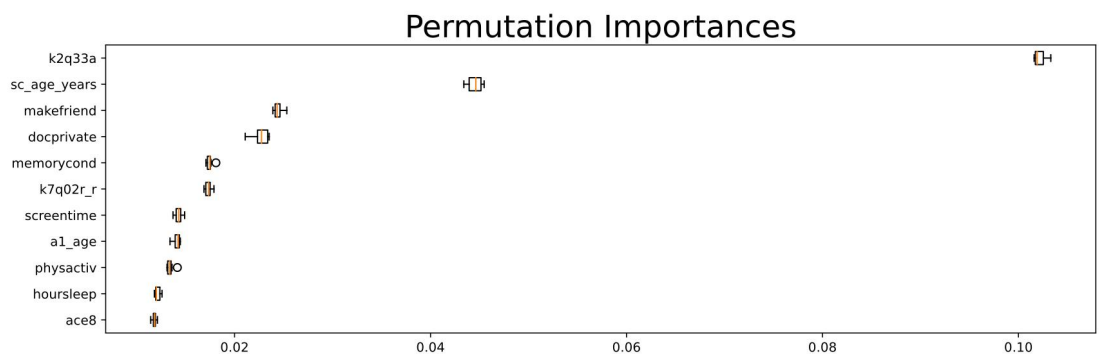

**Figure 4.** The feature importance dot plot shows the most important 11 variables for detection depression. The variable importance is measured by two classifiers: RandomForest and XGBoost. The result analysis reveals that compared to others, mental health and daily activities occupy a greater weight in judgment. This is also similar to the research of other scholars.

3.2. Basic Machine Learning and DNN Method

**Table 3.** Preliminary testing of the predictive ability of the basic model on the original dataset, including we deploy Decision Tree, XGBoost, logistic regression, SVM, and a simple multi-layer neural network. At the same time, we also extracted the test set as a sample balanced version, and the performance of XGBoost and neural network models was acceptable.

| Model | Imbalanced Testset | | Balanced Testset | |
|---|---|---|---|---|
| | ACC(%) | AUC | ACC(%) | AUC |
| Decision Tree | 93.66±0.05 | 0.73 | 70.92±0.05 | 0.71 |
| XGBoost | 95.71±0.05 | **0.73** | **72.71±0.05** | **0.73** |
| Logistic Regression | 95.58±0.05 | 0.69 | 69.35±0.05 | 0.693 |
| SVM | 94.82±0.05 | 0.53 | 51.91±0.05 | 0.52 |
| DNN | **95.81±0.05** | 0.702 | 71.42±0.05 | 0.704 |

3.3. Our Methods

We performed our experiments using our own proposed method. Through the scoring system, the data are scored, and our proposed adaptive method was used for model building. In the final experiment, we performed the sampling for five times. Each sampling we save the subsampled dataset as a new training dataset. Our test set is the same as the dataset from the previous underlying experiments. The obtained results are shown in the table.

**Table 4.** Experiment for different loss strategy in our proposed method. Focal Loss* is the improved version for data imbalance based on focal loss. The Boosted is the final aggregation model.

| Model | Train Set Volume | Loss Strategy | Imbalanced Testset | | Balanced Testset | |
|---|---|---|---|---|---|---|
| | | | ACC(%) | AUC | ACC(%) | AUC |
| 1 | 19687 | CrossEntropy | 89.63±0.05 | 0.92 | 82.67±0.05 | 0.93 |
| 2 | 20673 | CrossEntropy | 90.35±0.05 | 0.93 | 84.37±0.05 | 0.93 |
| 3 | 20132 | CrossEntropy | 91.12±0.05 | 0.94 | 84.53±0.05 | 0.94 |
| 4 | 20132 | Focal Loss | 88.97±0.05 | 0.91 | 83.57±0.05 | 0.93 |
| 5 | 20132 | GHM Loss | 0.899±0.05 | 0.92 | 85.72±0.05 | 0.93 |
| 6 | 20132 | Focal Loss* | **0.931±0.05** | **0.95** | **86.51±0.05** | **0.95** |
| Boosted | - | - | 90.50±0.05 | 0.94 | 86.38±0.05 | **0.95** |

In terms of results, although the accuracy of judgment decreased in the unbalanced dataset, the AUC results were significantly improved, which indicates that the processing power of the model in the unbalanced dataset has been improved. Because in the unbalanced data set, the proportion of negative samples is as high as 95%, and if all the outputs are negative, the accuracy will be as high as 95%, but this obviously does not mean that the model is a very poorly effective model. The test results on the data equilibrium test set show that the accuracy of all the tests is significantly improved compared with the basic model test, and the AUC of the experiment also improves to more than 0.92. The experimental results show that our work is meaningful.

At the same time, we extracted different datasets for training and updated the model to the latest version using the Boosted method. All the processes of data update, extraction, training and model update can be conducted online. This also proves that the whole process of the service cloud can be carried out without obstacles.

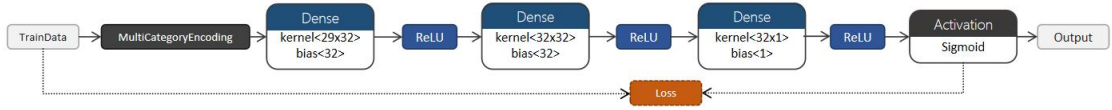

**Figure 5.** Different generation model construction of the proposed strategy. Our data processing process generates different training sets by sampling on the original dataset with data importance attributes, and fixes 5 training sets from the sampled training set for training. The models generated under different training sets may also vary to some extent, but the basic structure of the backbone network is shown in the figure. We also investigated the impact of custom loss on model performance.

In the final, we conducted comparative experiments between the baseline predictors and our proposed adaptive modeling approach, utilizing a synthetic training dataset processed with the CensusFilter data handling method. The training dataset was synthesized from two instances of CensusFilter sampling, totaling 40,289 data entries. The test set was chosen to be the same balanced dataset used previously.

**Table 5.** The result of method comparison under the data from CensusFilter generation.

| Model | Precision (±0.05) | Sensitive (±0.05) | Accuracy (±0.05) | F1 (±0.05) |
|---|---|---|---|---|
| XGBoost | 87.03 | 77.24 | 85.91 | 85.82 |
| SVM | 87.12 | 79.77 | 86.42 | 86.33 |
| DNN | 87.01 | 80.72 | 86.96 | 87.12 |
| Single Predictor | 87.95 | 81.95 | 87.28 | 87.18 |
| Boosted Predictor (AADA) | **88.25** | **82.50** | **87.92** | **87.80** |

The results were telling. Firstly, after applying our proposed CensusFilter data processing method, there was a notable improvement in the performance of both basic machine learning and deep neural network approaches, affirming the effectiveness of our data processing method. Secondly, it was observed that our AADA Predictor outperformed all others on the same dataset. Notably, some baseline methods exceeded the experimental data in **Table 5.** This discrepancy can be attributed to our previous experiments which utilized a dataset from a single sampling method, whereas the current experiment benefited from a training set synthesized from two rounds of sampling.

**4 Discussion**

4.1 Analysis and findings

The diagnosis of depression in adolescents has been convincingly demonstrated by many scholars to be a formidable task. Owing to the limitations of adolescents' perception and expression capabilities, their depression issues are often overlooked by their guardians.Amidst high depression rates and scarce child mental health specialists, primary care providers' comfort in diagnosing and treating depression is crucial. However, few pediatric programs offer relevant training, leaving many practitioners uneasy about treating these conditions[27]. Other research indicates that While depression diagnosis in children and adolescents mirrors adult criteria, developmental stages can alter symptom expression, and some youngsters may find it challenging to comprehend and communicate their emotional states[28]. Therefore, the objective of this study is to develop a depression risk model based on various objective factors to assist in determining the risk of adolescents developing depression. This carries significant implications for the diagnosis of adolescent depression and the prevention of such risks within a family setting.

Although numerous attempts have been made to employ computer-aided diagnosis of depression, the majority of studies still revolve around establishing automatic depression diagnosis systems based on clinical interview datasets[29–31]. Furthermore, a deep learning judgement framework for depression diagnosis, based on electroencephalogram data analysis, has achieved over 95% accuracy in clinical diagnoses[32]. With the advancement of multimodal technologies, methods for detecting depression through facial and voice recognition have been proposed. One study introduced an automatic audio-visual and multimodal emotion perception detection system for assessing the severity of depression, creating a data collection system and conducting experiments with the collected data[33]. Of course, there are also studies using medical imaging to assist in diagnosing depression, with MRI research uncovering potential biomarkers for depression. A review of these biomarkers has been published [34–36].

From the result of this study, preliminary data correlation analysis indicates a significant correlation between the occurrence of depression and other psychological disorders. For instance, the Pearson correlation coefficient between depression and anxiety is r=0.552. Experience suggests that the occurrence of psychological disorders such as depression, anxiety, and restlessness is often interrelated. Moreover, symptoms of depression and anxiety tend to exhibit more significant correlations with other diseases[37]. Adolescents with Autism Spectrum Disorder (ASD) may exhibit more awareness of their social communication challenges compared to younger children. Moreover, the occurrence of coexisting depression appears to align with higher functioning ASD forms[38]. Meanwhile, Frequent physiological symptoms may also indicate depression. It is worth mentioning that depression often occurs with painful symptoms, such as headache, back pain, joint pain[39]. Which are chosen as the variables in our study. We expect the designed system to evaluate the risk of mental health issues based on respondents' daily living conditions, lifestyle habits, and other objective factors. Therefore, we designed related experiments to validate our approach. The experiments demonstrate that, even when removing psychological issues such as anxiety, autism, and restlessness, our method still achieves an accuracy of 0.78±0.02 on our test set, with an AUC of 0.88±0.01. This further validates the effectiveness of our method in assessing depression risk. The volume of parameters required for input might make the system inconvenient to use. Research indicates that the length and design of a questionnaire can affect the quality of responses from participants[40]. In psychiatric consultations, an initial mental health evaluation could last over an hour, with specifics based on processes and methods such as the Mini International Neuropsychiatric Interview, Structured Clinical Interview, and Primary Care Evaluation of Mental Disorders[41]. To obtain better consultation quality in the diagnostic process, we plan to conduct a variable study in our future work. By continuously screening variables, we aim to build and test models, thereby finishing a variable study. This could help us employ fewer variables to establish a credible prediction model.

We conducted substantial analysis in the preparation stage of our study, particularly in basic data analysis. For instance, in the entirety of the data we used, we found that the incidence rate of depression in adolescent girls was 4.7%, higher than the 3.9% in boys. This finding has also been mentioned in the research of many scholars, suggesting that changes in gonadal steroids at puberty may lead to depression and anxiety, which explains the higher prevalence of depression in females [42].

Symptomatology is also a fundamental basis in the research of psychological disorders.In our research, we discovered that the census data hold plenty of usable information. In addition to psychological disorders such as anxiety and agitation, it is also possible to consider physiological issues like frequent headaches, toothaches, and other recurrent bodily ailments as research targets. Simultaneously, the data allow for the investigation of factors driving physical activity among adolescents. In the NSCH dataset, we find provided survey items including weekly amount of exercise and frequency of participation in sport teams. These data permit the examination of relationships between environmental factors, lifestyle habits, dietary quality, and adolescent willingness to engage in physical activity. This can provide guardians with more comprehensive information for improvement.

Upon analyzing all results, it is evident that certain variables have a significant impact on psychological issues such as depression, with strong correlations seen in areas of lifestyle and physical activity. In many studies related to adolescent depression, researchers have also established clear correlations between these factors and the onset of depression.Regular participation in physical exercise can help adolescents enhance their physical strength, reduce the risk of cardiovascular diseases and other chronic illnesses. Furthermore, it can also boost self-esteem and self-identity, and alleviate anxiety and stress [43].Research also indicates that regular physical exercise can alleviate various physiological and psychological issues in individuals with depression[44]. A study conducted in an East London community found evidence of a cross-sectional association between physical activity in adolescents and depressive symptoms. For each additional hour of exercise per week at a basic level of physical activity, the odds of depressive symptoms decrease by about 8%[45]. In our research, we individually tested variables related to the 'Living/Sport' category and found that lifestyle habits and physical activity not only have a significant correlation with the risk of depression, but also impact the accuracy of the predictive model in independent validation experiments. Specifically, discarding entries related to physical activity reduces the accuracy of the predictive model.

4.2 Limitation

There are still some limitations in this study. All our current data are sourced from the cross-sectional data of the NSCH census spanning 2020-2022. These data do not permit further time-series tracking. Consequently, our study cannot ascertain the causal relationship between variables and outcomes; we can only predict the outcomes after the variables have been processed through the deep neural network. Moreover, we cannot validate whether the variables that need improvement, as suggested by the system, could result in meaningful depression risk reduction in real-world clinical outcomes. As our data are derived from the U.S. census, the findings based on this dataset might only be applicable to local populations, same ethnicity groups, or those with similar cultural and temporal backgrounds. As such, it may not represent adolescent samples from all global and temporal backgrounds. However, a distinctive feature of our designed system is that it obviates the need for laborious feature engineering, enabling rapid model generation and analysis with new data.

For risk prediction and determination, the system currently provides only two types of classification results: normal and at-risk of depression. However, we aim to enhance the system with a more comprehensive scoring mechanism, including quantitative assessment of risk. This will offer risk prevention advice for individuals or guardians, emphasizing the need for monitoring and improvement of specific variables.

Moreover, the census contains some items with missing entries, and certain model computations may not accept these missing values. We have uniformly filled in these missing values as -1, but this might affect data analysis and the accuracy of model judgment. Although deleting samples with missing values is a direct and efficient method, our practical experiment statistics found that missing values account for more than 80% of all data. Using the direct deletion method would result in the loss of a considerable amount of sample information. Other methods of dealing with missing values, such as mean imputation, regression imputation, maximum likelihood estimation, and multiple imputation[46], could be attempted in experiments to further enhance the accuracy of the model.

**5. Conclusion**

In our work, we proposed a method for psychological analysis based on big data, which is founded on traditional data processing and novel neural networks. We selected the American adolescent mental health survey as the subject of our study to investigate the hypothesis of our proposed method. The experiments validated the effectiveness of our approach by comparing it on the dataset with traditional machine learning algorithms and simple deep learning networks. The results demonstrated that our proposed methodology significantly outperforms traditional machine learning, general deep learning, and other medical prediction methods proposed by others. It can be used for auxiliary diagnosis of adolescent psychological issues in multidimensional questionnaire diagnosis.

However, there are still deficiencies. The current overall prediction accuracy has not yet reached a high level, and there is still bias in judgment. In the future, we hope to further enhance the judgment ability of the model using correlation analysis methods. Also, the application of artificial intelligence in medical diagnosis is still in the research stage, and trustworthy AI technology is being advocated. We look forward to employing causal inference methods to enhance the credibility of the data in our future work.